\def\year{2021}\relax
\documentclass[letterpaper]{article} 
\usepackage{aaai21}  
\usepackage{times}  
\usepackage{helvet} 
\usepackage{courier}  
\usepackage[hyphens]{url}  
\usepackage{graphicx} 
\usepackage{natbib}  
\usepackage{caption} 
\usepackage{multirow}
\usepackage{amsmath}
\usepackage{subfigure}
\usepackage[noend]{algpseudocode}
\usepackage{algorithmicx,algorithm}
\usepackage{array}
\def\eg{\emph{e.g.~}} 
\def\ie{\emph{i.e.~}}

\title{Peer Collaborative Learning for Online Knowledge Distillation}
\author{
    Guile Wu, Shaogang Gong\\
}
\affiliations{
    Queen Mary University of London\\
    guile.wu@qmul.ac.uk, s.gong@qmul.ac.uk

}

\begin{document}

\maketitle

\begin{abstract}
   Traditional knowledge distillation uses a two-stage training strategy to transfer knowledge
   from a high-capacity teacher model to a compact student model, which relies heavily on the
   pre-trained teacher.
   Recent online knowledge distillation alleviates this limitation by collaborative learning,
   mutual learning and online ensembling, following a one-stage end-to-end training fashion.
   However, collaborative learning and mutual learning fail to construct an online high-capacity teacher,
   whilst online ensembling ignores the collaboration among branches and its logit summation
   impedes the further optimisation of the ensemble teacher.
   In this work, we propose a novel Peer Collaborative Learning method
   for online knowledge distillation,
   which integrates online ensembling and network collaboration into
   a unified framework.
   Specifically, given a target network,
   we construct a multi-branch network for training, in which each branch is called a peer.
   We perform random augmentation multiple times on the inputs to peers
   and assemble feature representations outputted from peers with an additional classifier as the peer ensemble teacher.
   This helps to transfer knowledge from a high-capacity teacher to peers,
   and in turn further optimises the ensemble teacher.
   Meanwhile, we employ the temporal mean model of each peer as the peer mean teacher
   to collaboratively transfer knowledge among peers, which
   helps each peer to learn richer knowledge
   and facilitates to optimise a more stable model
   with better generalisation.
   Extensive experiments on CIFAR-10, CIFAR-100 and ImageNet show that the proposed method
   significantly improves the generalisation of various backbone networks
   and outperforms the state-of-the-art methods.
\end{abstract}

\section{Introduction}

Deep learning has achieved incredible success in many computer vision tasks in recent years.
Whilst many studies focus on developing deeper and/or wider
networks for improving the performance~\cite{he2016deep,zagoruyko2016wide,xie2017aggregated},
these cumbersome networks require more computational resources,
which hinders their deployments in resource-limited scenarios.
To alleviate this problem, knowledge distillation is developed to transfer knowledge from a
stronger teacher~\cite{hinton2015distilling}
or an online ensemble~\cite{lan2018knowledge}
to a student model, which is more suitable for deployment.

Traditionally, knowledge distillation (KD) requires to pre-train a high-capacity teacher model
in the first stage, and then transfer the knowledge of the teacher to a smaller student model in the
second stage~\cite{hinton2015distilling,romero2015fitnets,phuong2019towards}.
Via aligning soft predictions~\cite{hinton2015distilling} or
feature representations~\cite{romero2015fitnets} between the teacher and the student,
a student model usually significantly reduces the model complexity for deployment
but still achieves competitive accuracy as the teacher model.
However, since the teacher and the student are trained in two separate stages,
this traditional strategy usually requires more training time and computational cost.

\begin{figure*}[t]
\begin{center}
   \includegraphics[width=0.85\linewidth]{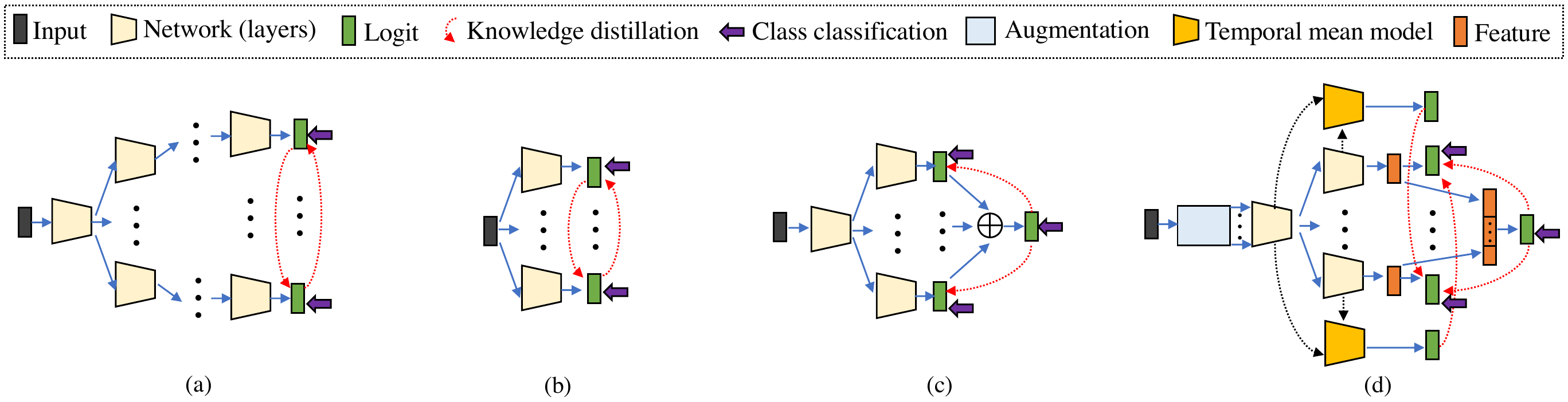}
\end{center}
   \caption{Comparing four online knowledge distillation mechanisms:
   (a) Collaborative learning.
   (b) Mutual learning.
   (c) Online ensembling.
   (d) Peer collaborative learning (Proposed).
   Our method integrates two types of peer collaborations (\ie peer ensemble teacher and peer mean teacher)
   into a unified framework to improve the quality of online knowledge distillation.}
   \label{fig:illustration}
\end{figure*}

Recent online knowledge distillation
~\cite{lan2018knowledge,zhang2018deep,chen2020online} alleviates this limitation by
directly optimising a target network, following a one-stage end-to-end training fashion.
Instead of pre-training a high-capacity teacher,
online knowledge distillation typically integrates the teacher into the student using
a hierarchical network with shared intermediate-level representations~\cite{song2018collaborative}
(Fig.~\ref{fig:illustration}(a)),
multiple parallel networks~\cite{zhang2018deep}(Fig.~\ref{fig:illustration}(b)),
or a multi-branch network with online ensembling~\cite{lan2018knowledge}
(Fig.~\ref{fig:illustration}(c)).
Although these methods have shown their superiority over the traditional counterparts,
collaborative learning and mutual learning fail to construct an online high-capacity teacher
to facilitate the optimisation of the student,
whilst online ensembling ignores the collaboration among branches and its logit summation
impedes the further optimisation of the ensemble teacher.

In this work, we propose a novel Peer Collaborative Learning (PCL) method for online knowledge distillation.
As shown in Fig.~\ref{fig:illustration}(d), we integrate online ensembling and network collaboration into
a unified framework to take full advantage of them for improving the quality of online knowledge distillation.
Specifically, in training, we construct a multi-branch network
by adding auxiliary branches (high-level layers) to a given target network.
We call each branch ``{\em a peer}'' and design two types of online teachers for peer collaborative learning to improve
the generalisation of a target network.
The first teacher, \emph{peer ensemble teacher}, is an online high-capacity model, which helps to distil knowledge from a stronger ensemble
teacher to each peer, and in turn further improves the ensemble teacher.
Instead of using peer logit summation to construct the ensemble teacher~\cite{lan2018knowledge},
we perform random augmentation multiple times on the inputs to peers
and then assemble feature representations outputted from peers with an additional classifier as the peer ensemble teacher.
This design helps to learn discriminative information among feature representations of peers and
facilitates to assemble a stronger teacher for online knowledge distillation.
Furthermore, to generate a more stable model with better generalisation,
we use the second teacher, \emph{peer mean teacher},
to collaboratively distil knowledge among peers.
Instead of directly distilling knowledge among peers using mutual learning~\cite{zhang2018deep},
we utilise the temporal mean model of each peer
to construct the peer mean teacher which can produce more stable predictions.
As a result, this design helps each peer to learn richer knowledge
and facilitates to optimise a more stable model with better generalisation for deployment.
In testing, we use a temporal mean model of a peer for deployment,
which has the same number of parameters as the given target network,
so there is no extra inference cost for deployment.
Besides, the outputted feature representations from peer mean teachers plus the additional classifier
can form a high-capacity ensemble for deployment to get better performance
in the scenarios where computational cost is less constrained.

Our {\bf contributions} are:
{\bf (I)} We propose a novel Peer Collaborative Learning method for online knowledge distillation,
which integrates online ensembling and network collaboration into a unified framework;
{\bf (II)} We construct a peer ensemble teacher via performing random augmentation multiple times on the inputs
to peers and assembling feature representations outputted from peers with an additional classifier.
This helps to simultaneously optimise peers and the ensemble teacher for online knowledge distillation.
{\bf (III)} We utilise the temporal mean model of each peer to construct the peer mean teacher
for peer collaborative distillation, resulting in a more stable model with better generalisation;
{\bf (IV)} Extensive experiments on CIFAR-10~\cite{krizhevsky2009learning},
CIFAR-100~\cite{krizhevsky2009learning} and ImageNet~\cite{russakovsky2015imagenet}
using a variety of backbone networks
show that the proposed method significantly improves the
generalisation of the backbone networks and outperforms the state-of-the-art alternative methods.

\begin{figure*}
\begin{center}
   \includegraphics[width=0.61\linewidth]{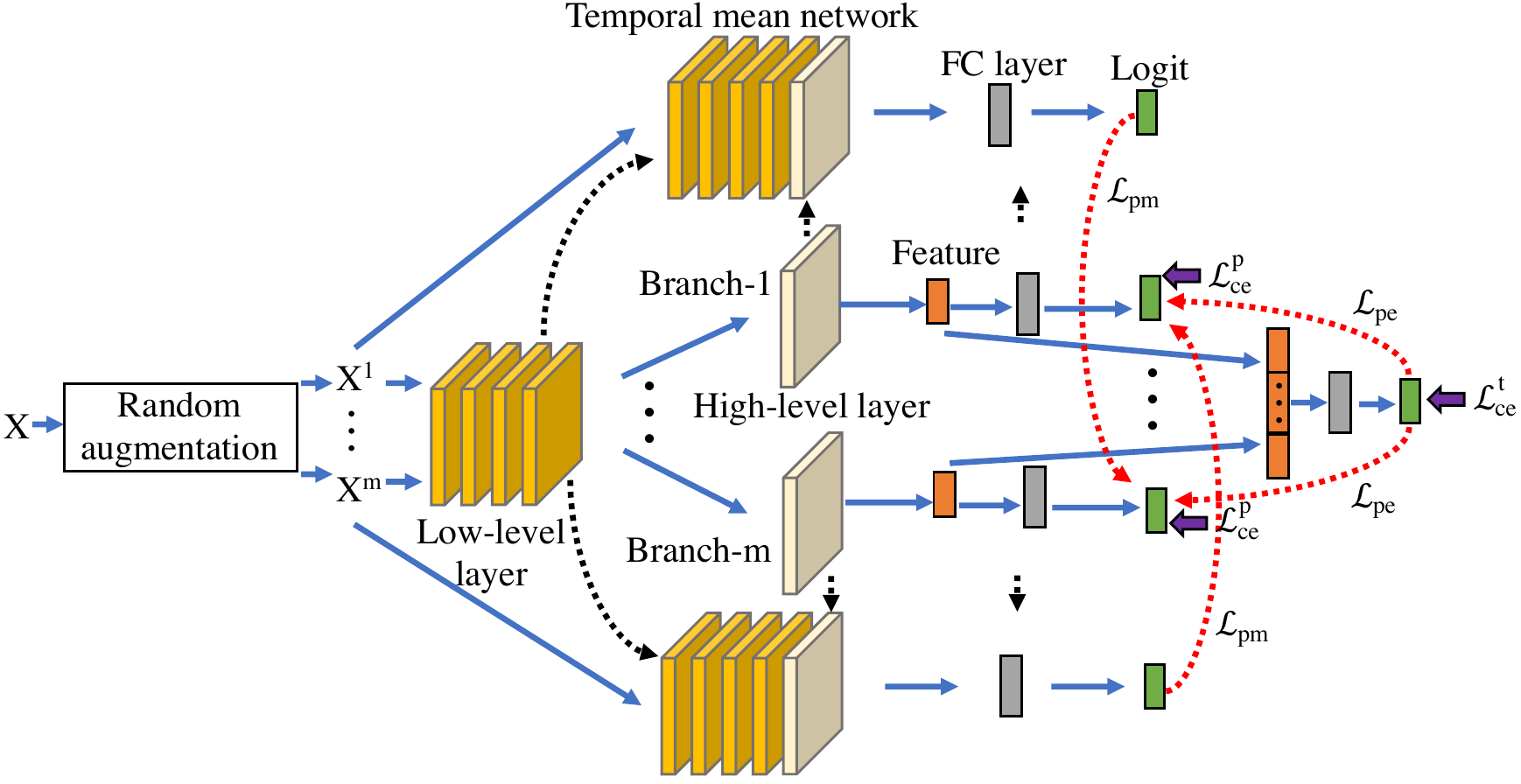}
\end{center}
   \caption{An overview of Peer Collaborative Learning (PCL) for online knowledge distillation.
   }
   \label{fig:overview}
\end{figure*}

\section{Related Work}
{\bf Traditional Knowledge Distillation}~\cite{hinton2015distilling}
is one of the most effective solutions to compress a cumbersome model
or an ensemble of models into a smaller model for deployment.
In~\cite{hinton2015distilling}, \citeauthor{hinton2015distilling} propose to
distil the knowledge from a high-capacity teacher model to a compact student model,
which is accomplished by aligning soft output predictions between the teacher and the student.
In recent years, many promising methods have been designed to
transfer various ``knowledge'', such as
intermediate representations~\cite{romero2015fitnets}, flow between layers~\cite{yim2017gift},
attention maps~\cite{zagoruyko2016paying}, structural relations~\cite{park2019relational} and
activation similarity~\cite{tung2019similarity},
to facilitate the optimisation process of distillation.
Although these methods have shown competitive performance for compressing a model,
they typically follow a two-stage training solution by pre-training a high-capacity teacher model
for transferring knowledge to a compact student model,
which requires more training time and computational cost.

{\noindent\bf Online Knowledge Distillation}~\cite{lan2018knowledge,chen2020online,zhang2018deep}
proposes to directly optimise a target network via distilling knowledge among multiple networks or branches
without pre-training a high-capacity teacher, which follows a one-stage end-to-end training strategy.
Since online KD directly optimises a target network,
there is no need to store or download a teacher model, which saves time and computational cost.
In~\cite{song2018collaborative}, \citeauthor{song2018collaborative} propose to distil knowledge among
multiple classifier heads of a hierarchical network for improving the generalisation
of a target network.
In~\cite{zhang2018deep}, \citeauthor{zhang2018deep} introduce a mutual learning solution to distil knowledge
among multiple parallel networks with the same input.
Although these methods help to improve the generalisation of the target network,
they only distil limited knowledge among parallel networks or heads and fail to construct a stronger
online teacher to further improve the student.
In~\cite{guo2020online}, \citeauthor{guo2020online} employ multiple parallel networks and aggregate logits from
all student networks based on the cross-entropy loss to generate soft targets for online distillation.
More similar to our work, \citeauthor{lan2018knowledge}~\cite{lan2018knowledge} use a multi-branch network
and assemble logits from multiple branches (students) as the teacher to improve the generalisation
of each student.
However, logit aggregation impedes the further optimisation of the ensemble teacher
and online ensembling ignores the collaboration among branches, resulting in suboptimal performance.
In~\cite{kim2019feature}, \citeauthor{kim2019feature} integrate feature representations of multiple branches
into the online ensembling, but their method requires more convolutional operations for the feature
fusion and also fails to exploit the collaboration among branches.
To address these limitations, in our work:
(1) we assemble feature representations from peers with an additional classifier as the \emph{peer ensemble teacher},
which helps to distil knowledge from an online high-capacity teacher to each peer (student) and in turn further optimises the teacher;
(2) we exploit the temporal mean model of each peer as the \emph{peer mean teacher}
to distil knowledge among peers, which helps each peer to learn richer knowledge
and facilitates to optimise a more stable model.
Integrating these two teachers into a unified framework significantly improves the generalisation
of each peer and the ensemble, resulting in better performance.

{\noindent\bf Neural Network Ensembling} is a simple and effective solution for improving the generalisation
performance of a model~\cite{hansen1990neural,zhou2002ensembling,moghimi2016boosted}.
Although this can usually bring better performance, training multiple neural networks to create an ensemble
requires significantly more training time and computational cost.
The recent trend in neural network ensembling focuses on training a single model and exploiting different training
phases of a model as an ensemble.
In~\cite{huang2017snapshot}, \citeauthor{huang2017snapshot} force the model to visit multiple local minima and
use the corresponding models as the snapshots for neural network ensembling.
In~\cite{laine2017temporal}, \citeauthor{laine2017temporal} propose to use temporal ensembling of network predictions
over multiple training epochs as the teacher to facilitate the optimisation of the current model for semi-supervised learning.
Our work differs from these works in that we use feature representations of peers from a multi-branch network with an additional classifier as the
ensemble teacher for online knowledge distillation, rather than using the network predictions from different phases
or generating multiple networks for ensembling.
In our method, the peer mean teacher shares the merit of the traditional Mean
Teacher~\cite{tarvainen2017mean}.
In~\cite{tarvainen2017mean}, network weights over previous training epochs are aggregated as a teacher
to minimise the distance of predictions between the student and the teacher as the consistency
regularisation for semi-supervised learning.
In contrast, our method uses the shared low-level layers and multiple separated high-level layers to construct multiple
peer mean teachers for aligning soft prediction distributions between the peer and its counterparts’ mean teacher,
resulting in a more stable model for improving the quality of
online knowledge distillation.

\section{Peer Collaborative Learning}

\subsection{Approach Overview}
The overview of the proposed Peer Collaborative Learning (PCL) is depicted in Fig.~\ref{fig:overview}.
We employ an $m$-branch network for model training and call each branch ``\emph{a peer}''.
Since the low-level layers across different branches usually contain similar low-level features
regarding minor details of images, sharing them enables to reduce the training cost and improve the collaboration
among peers~\cite{lan2018knowledge}.
We therefore share the low-level layers and separate the high-level layers in the $m$-branch network.

As shown in Fig.~\ref{fig:overview}, to facilitate online knowledge distillation,
we assemble the feature representation of peers with an additional classifier as the \emph{peer ensemble teacher}
and use the temporal mean model of each peer as the \emph{peer mean teacher}.
The training optimisation objective of PCL contains three components:
The first component is the standard cross-entropy loss for classification
of the peers ($\mathcal{L}_{ce}^p$) and the peer ensemble teacher ($\mathcal{L}_{ce}^t$);
The second component is the peer ensemble teacher distillation loss $\mathcal{L}_{pe}$ for transferring
knowledge from a stronger teacher to a student, which in turn further improves the ensemble teacher;
The third component is the peer mean teacher distillation loss $\mathcal{L}_{pm}$ for collaboratively
distilling knowledge among peers.
Thus, the overall objective $\mathcal{L}$ is formulated as:
\begin{equation}
\label{eq:loss_all}
\mathcal{L} = \mathcal{L}_{ce}^p + \mathcal{L}_{ce}^t + \mathcal{L}_{pe} + \mathcal{L}_{pm}.
\end{equation}

In testing, we use a temporal mean model of a peer for deployment,
which has the same number of parameters as the backbone network,
so there is no extra inference cost for deployment.
In the scenarios where computational cost is less constrained,
feature representations from peer mean teachers plus the additional classifier
can form an ensemble model for deployment to get better performance.

\subsection{Peer Ensemble Teacher}
\noindent{\bf Input Augmentation for Peers.}
Suppose there are $n$ samples in a training dataset $\{(x_i, y_i)\}_{i=1}^n$,
where $x_i$ is the $i$-$th$ input sample, $y_i\in\{1,2,...,C\}$ is the corresponding label, and
$C$ is the number of classes in the dataset ($C{\leq}n$).
Existing multi-branch online distillation methods~\cite{lan2018knowledge,chen2020online}
directly use $x_i$ (applying random augmentation once) as the input to all the branches, which
causes the homogenisation among peers and decreases the generalisation of the network.
To alleviate this problem, we perform random augmentation $m$ times to $x_i$ to
generate $m$ counterparts of $x_i$ (\ie $\{x_i^1, x_i^2, ..., x_i^m\}$,
and use each counterpart as the input to each peer.
This stochastic augmentation fashion is similar to~\cite{laine2017temporal},
but we perform it multiple times to assemble discriminative features as the ensemble
teacher in a multi-branch network, rather than to generate two predictions for
consistency regularisation or distillation.

\noindent{\bf Online Ensembling.}
To construct a stronger online teacher for online knowledge distillation,
logits from multiple networks/branches are usually aggregated (w/ or w/o attention gates)
\cite{lan2018knowledge,chen2020online}.
However, this impedes the ensemble teacher from further optimising the ensemble model and
ignores the discriminative information among feature representations of peers,
which might lead to a suboptimal solution since the logit summation is not further learned.
In our work, we concatenate the features outputted from peers and use an additional fully connected layer
for classification to construct a learnable stronger online teacher.
Thus, the multi-class classification is performed for both the peers and the ensemble teacher as:
\begin{equation}
\label{eq:loss_ce_p}
\mathcal{L}_{ce}^p=-\sum_{j=1}^m\sum_{c=1}^C{y_c}log\frac{exp(z_{j,c}^p)}{\sum_{k=1}^Cexp(z_{j,k}^p)},
\end{equation}
\begin{equation}
\label{eq:loss_ce_t}
\mathcal{L}_{ce}^t=-\sum_{c=1}^Cy_clog\frac{exp(z_{c}^t)}{\sum_{k=1}^Cexp(z_{k}^t)},
\end{equation}
where $z_{j,c}^p$ is the output logit from the last fully connected layer of the $j$-th peer
over a class $c$, $y_c$ is the ground-truth label indicator,
$z_c^t$ is the output logit from the fully connected layer of the peer ensemble teacher over a class $c$.

Then, to transfer knowledge from the ensemble teacher to each peer, we compute the soft prediction of the $j$-$th$
peer and the ensemble teacher as:
\begin{equation}
\label{eq:soft_peer_ensemble}
p_{j,c}^p = \frac{exp(z_{j,c}^p/T)}{\sum_{k=1}^Cexp(z_{j,k}^p/T)},\
p_{c}^{t} = \frac{exp(z_{c}^t/T)}{\sum_{k=1}^Cexp(z_{k}^t/T)},
\end{equation}
where $T$ is a temperature parameter~\cite{hinton2015distilling},
$p_{j,c}^p$ is the soft prediction of the $j$-$th$ peer over a class $c$,
and $p_{c}^t$ is the soft prediction of the ensemble teacher over a class $c$.
Using Kullback Leibler (KL) divergence, the peer ensemble distillation loss $\mathcal{L}_{pe}$
is formulated as:
\begin{equation}
\label{eq:loss_pe}
\mathcal{L}_{pe}=\omega(e){\cdot}T^2\sum_{j=1}^m\sum_{c=1}^C{p_{c}^{t}{\cdot}log\frac{p_{c}^{t}}{p_{j,c}^p}},
\end{equation}
where $T^2$ ensures contributions of ground-truth and teacher probability distributions keep roughly unchanged
\cite{hinton2015distilling}, $e$ is the current training epoch,
$\omega(\cdot)$ is a weight ramp-up function~\cite{laine2017temporal} stabilises model training,
which is defined as:
\begin{equation}
\label{eq:loss_ramp}
\omega(e)= 
\left\{
\begin{aligned}
&\lambda{\cdot}exp(-5*(1-\frac{e}{\alpha})^2)&,e\leq\alpha;\\
&\lambda  &,e>\alpha;
\end{aligned}\right.
\end{equation}
where $\alpha$ is the epoch threshold for the ramp-up function
and $\lambda$ is a parameter weighting the gradient magnitude.

\noindent{\bf Remarks.}
The proposed peer ensemble teacher differs from existing feature fusion
\cite{hou2017dualnet,kim2019feature,chen2017person} in that
we construct an online high-capacity teacher model by performing random augmentation multiple times on
the input to peers and assembling feature representations from peers
of a multi-branch network with an additional classifier,
without using additional convolutional operations or multiple networks.
This helps to effectively distil knowledge from a stronger ensemble teacher to each peer,
and in turn further improves the ensemble teacher.

\subsection{Peer Mean Teacher}
Online ensembling helps to construct a stronger teacher for online knowledge distillation,
but it ignores the collaboration among peers.
On the other hand, mutual learning~\cite{zhang2018deep} and collaborative
learning~\cite{song2018collaborative} benefit from mutual distillation among networks or heads,
but they fail to construct a high-capacity teacher for online distillation.
In our work, we further use peer mutual distillation for improving collaboration among peers.
Instead of directly distilling knowledge among peers,
we use the temporal mean model~\cite{tarvainen2017mean} of each peer as
the peer mean teacher for peer collaborative distillation.
We denote the weights of the shared low-level layers as $\theta_{l}$
and the weights of the separated high-level layers of the $j$-$th$ peer as $\theta_{h,j}$.
At the $g$-$th$ global training step
\footnote{Here, $g=e{\cdot}$Batch$_{num}$ + Batch$_{ind}$,
where Batch$_{num}$ is the total number of training mini-batches in each epoch and Batch$_{ind}$
is the index of the current mini-batch.},
the $j$-$th$ peer mean teacher $\{\theta_{l,g}^t, \theta_{h,j,g}^t\}$ is formulated as:
\begin{equation}
\label{eq:mean_teacher}
\left\{
\begin{aligned}
\theta_{l,g}^t &= \phi(g){\cdot}\theta_{l,g-1}^t + (1-\phi(g)){\cdot}\theta_{l,g},\\
\theta_{h,j,g}^t &= \phi(g){\cdot}\theta_{h,j,g-1}^t + (1-\phi(g)){\cdot}\theta_{h,j,g},
\end{aligned}\right.
\end{equation}
where $\theta_{l,g}^t$ are the weights of the shared low-level layers of the peer mean teachers,
$\theta_{h,j,g}^t$ are the weights of the separated high-level layers of the $j$-$th$ peer mean teacher,
$\phi(g)$ is a smoothing coefficient function defined as:
\begin{equation}
\label{eq:smoothing}
\phi(g)=min(1-\frac{1}{g},\beta),
\end{equation}
where $\beta$ is the smoothing coefficient hyper-parameter.
Note that, the additional classifier of the peer ensemble teacher is also aggregated
for the ensemble deployment.
We compute the soft prediction $p_{j,c}^{mt}$ of the $j$-$th$ mean teacher over a class $c$
as Eq.~(\ref{eq:soft_peer_ensemble}) with the output logit $z_{l,c}^{mt}$ of this mean teacher.
Thus, the peer mean teacher distillation loss $\mathcal{L}_{pm}$ is formulated as:
\begin{equation}
\label{eq:loss_pm}
\mathcal{L}_{pm}=\omega(e){\cdot}\frac{{T^2}}{m-1}\sum_{j=1}^m\sum_{l=1,l{\neq}j}^{m}\sum_{c=1}^C
{p_{l,c}^{mt}{\cdot}log\frac{p_{l,c}^{mt}}{p_{j,c}^p}}.
\end{equation}

\noindent{\bf Remarks.}
The traditional mean teacher is used for semi-supervised/unsupervised learning
\cite{tarvainen2017mean,mittal2019semi,ge2020mutual},
which mainly enforces the distance between the model predictions to be close.
In contrast, we employ the temporal mean model of each peer in a multi-branch network as the peer mean teacher,
and use it for peer collaborative distillation by aligning the soft distributions between the peer and its
counterparts' mean teacher.
Compared with mutual distillation among peers~\cite{zhang2018deep},
averaging model weights temporally over training epochs enables the peer mean teacher to 
stabilise soft predictions for improving peer collaboration
and generating a more stable network.

\noindent{\bf Summary.}
As shown in Algorithm~\ref{algorithm:algorithm},
PCL follows a one-stage training fashion
without pre-training a high-capacity teacher.
In test, we use a single peer mean model as the target model (PCL) without adding extra inference cost.
Besides, the ensemble of peer mean teachers (with the additional classifier) can be used as a high-capacity ensemble (PCL-E).

\begin{algorithm}[t]
\caption{Peer Collaborative Learning for Online KD.}

{\bf Input:} Training data $\{(x_i, y_i)\}_{i=1}^n$.\\
{\bf Output:} A trained target model $\{\theta_{l}^t, \theta_{h,1}^t\}$,

~~~~~~~~~~~~~~and a trained ensemble model $\{\theta_{l}^t, \theta_{h,j}^t\}_{j=1}^m$.

\begin{algorithmic}[1]

\State /* \emph{Training} */

\State {\bf Initialisation:} Randomly initialise model parameters

\For{$e=0{\rightarrow}Epoch_{max}$}~~~~~/* \emph{Mini-Batch SGD} */

   \State Randomly transform $x_i$ to get counterparts $\{x_i\}_{j=1}^m$
   \State Compute features and logits of peers
   \State Assembling features as the peer ensemble teacher
   \State Compute the logits of the teachers
   \State Compute peer classification loss $\mathcal L_{ce}^p$ (Eq.(\ref{eq:loss_ce_p}))
   \State Compute ensemble classification loss $\mathcal L_{ce}^t$ (Eq.(\ref{eq:loss_ce_t}))
   \State Compute peer ensemble distillation loss $\mathcal L_{pe}$(Eq.(\ref{eq:loss_pe}))
   \State Compute mean teacher distillation loss $\mathcal L_{pm}$(Eq.(\ref{eq:loss_pm}))
   \State Update peer models with Eq.(\ref{eq:loss_all})
   \State Update peer mean teachers with Eq.(\ref{eq:mean_teacher})
\EndFor
\State {\bf end for}
\State /* \emph{Testing} */
\State Deploy with a single target model $\{\theta_{l}^t, \theta_{h,1}^t\}$
\State Deploy with an ensemble model $\{\theta_{l}^t, \theta_{h,j}^t\}_{j=1}^m$
\end{algorithmic}
\label{algorithm:algorithm}
\end{algorithm}

\section{Experiment}
\noindent{\bf Datasets.}
We used three image classification benchmarks for evaluation:
(1) \emph{CIFAR-10}~\cite{krizhevsky2009learning} contains 60000 images in 10 classes,
with 5000 training images and 1000 test images per class.
(2) \emph{CIFAR-100}~\cite{krizhevsky2009learning} consists of 60000 images in 100 classes,
with 500 training images and 100 test images per class.
(3) \emph{ImageNet ILSVRC 2012}~\cite{russakovsky2015imagenet} contains 1.2 million training images
and 50000 validation images in 1000 classes.

\noindent{\bf Implementation Details.}
To verify the effectiveness of our method, we used a variety of backbone networks,
including ResNet~\cite{he2016deep}, VGG~\cite{simonyan2015very}, DenseNet~\cite{huang2017densely},
WRN~\cite{zagoruyko2016wide}, and ResNeXt~\cite{xie2017aggregated}.
Following~\cite{lan2018knowledge}, the last block of each backbone network
was separated from the parameter sharing (on ImageNet, the last two blocks were separated),
while the other low-level layers were shared.
We set $m$=$3$ peers in the multi-branch architecture.
We used standard random crop and horizontal flip for the random augmentation in training,
and did not use random augmentation in testing.
We used SGD as the optimiser with Nesterov momentum $0.9$ and weight decay $5e$-$4$.
We trained the network for $Epoch_{max}$=$300$ epochs on CIFAR-10/100 and $90$ epochs on ImageNet.
We set the initial learning rate to 0.1, which decayed to \{0.01, 0.001\}
at \{150, 225\} epochs on CIFAR-10/100 and at \{30, 60\} epochs on ImageNet.
We set the batch size to 128,
the temperature $T$=$3$, $\alpha$=$80$ for ramp-up weighting,
$\beta$=$0.999$ to learn temporal mean models,
$\lambda$=$1.0$ for CIFAR-10/100 and $\lambda$=$0.1$ for ImageNet.
By default, in PCL, we used the first branch as the target network.
Our models were implemented with Python 3.6 and PyTorch 0.4, and trained on TESLA V100 GPU (32GB).

\noindent{\bf Evaluation Metrics.}
We used the top-1 classification error rate (\%) and
reported the average results with the standard deviation over 3 runs.

\begin{table*}[t]
   \begin{center}
   \small
      \begin{tabular}{c|c|p{0.15\columnwidth}p{0.15\columnwidth}p{0.15\columnwidth}p{0.15\columnwidth}p{0.15\columnwidth}
      p{0.16\columnwidth}|p{0.15\columnwidth}p{0.16\columnwidth}}
         \hline
         \multirow{1}{*}{Dataset}&\multirow{1}{*}{Network}&\centering{DML}&\centering{CL}&\centering{ONE}&\centering{FFL-S}
         &\centering{OKDDip}&\centering{KDCL}&\centering{Baseline}&PCL(ours)\\
         \hline\hline
         \multirow{6}{*}{C10}
         &ResNet32&6.06$\pm$0.07&5.98$\pm$0.28&5.80$\pm$0.12&5.99$\pm$0.11&5.83$\pm$0.15&5.99$\pm$0.08&6.74$\pm$0.15&{\bf 5.67$\pm$0.12}\\
         &ResNet110&5.47$\pm$0.25&4.81$\pm$0.11&4.84$\pm$0.30&5.28$\pm$0.06&4.86$\pm$0.10&4.89$\pm$0.16&5.31$\pm$0.10&{\bf 4.47$\pm$0.16}\\
         &VGG16&5.87$\pm$0.07&5.86$\pm$0.15&5.86$\pm$0.23&6.78$\pm$0.08&6.02$\pm$0.06&5.91$\pm$0.12&6.04$\pm$0.13&{\bf 5.26$\pm$0.02}\\
         &DenseNet40-12&6.41$\pm$0.26&6.95$\pm$0.25&6.92$\pm$0.21&6.72$\pm$0.16&7.36$\pm$0.22&6.13$\pm$0.08&6.81$\pm$0.02&{\bf 5.87$\pm$0.13}\\
         &WRN20-8&4.80$\pm$0.13&5.41$\pm$0.08&5.30$\pm$0.14&5.28$\pm$0.13&5.17$\pm$0.15&4.73$\pm$0.16&5.32$\pm$0.01&{\bf 4.58$\pm$0.04}\\
         &ResNeXt29&4.46$\pm$0.16&4.45$\pm$0.18&4.27$\pm$0.10&4.67$\pm$0.04&4.34$\pm$0.02&4.02$\pm$0.27&4.72$\pm$0.03&{\bf 3.93$\pm$0.09}\\
         \hline
         \multirow{6}{*}{C100}
         &ResNet32&26.32$\pm$0.14&27.67$\pm$0.46&26.21$\pm$0.41&27.82$\pm$0.11&26.75$\pm$0.38&26.24$\pm$0.34&28.72$\pm$0.19&{\bf 25.86$\pm$0.16}\\
         &ResNet110&22.14$\pm$0.50&21.17$\pm$0.58&21.60$\pm$0.36&22.78$\pm$0.41&21.46$\pm$0.26&21.72$\pm$0.32&23.79$\pm$0.57&{\bf 20.02$\pm$0.55}\\
         &VGG16&24.48$\pm$0.10&25.67$\pm$0.08&25.63$\pm$0.39&29.13$\pm$0.99&25.32$\pm$0.05&24.33$\pm$0.22&25.68$\pm$0.19&{\bf 23.11$\pm$0.25}\\
         &DenseNet40-12&26.94$\pm$0.31&28.55$\pm$0.34&28.40$\pm$0.38&28.75$\pm$0.35&28.77$\pm$0.14&27.48$\pm$0.42&28.97$\pm$0.15&{\bf 26.91$\pm$0.16}\\
         &WRN20-8&20.23$\pm$0.07&20.60$\pm$0.12&20.90$\pm$0.39&21.78$\pm$0.14&21.17$\pm$0.06&20.63$\pm$0.30&21.97$\pm$0.40&{\bf 19.49$\pm$0.49}\\
         &ResNeXt29&18.94$\pm$0.01&18.41$\pm$0.07&18.60$\pm$0.25&20.18$\pm$0.33&18.50$\pm$0.11&18.64$\pm$0.18&20.57$\pm$0.43&{\bf 17.38$\pm$0.23}\\
         \hline
         \multirow{1}{*}{ImgNet}
         &ResNet18&30.18$\pm$0.08&29.96$\pm$0.05&29.82$\pm$0.13$^\dagger$&31.15$\pm$0.07&30.07$\pm$0.06&30.40$\pm$0.05&30.49$\pm$0.14&{\bf 29.58$\pm$0.13}\\
        \hline
      \end{tabular}
   \end{center}
   \caption{Comparisons with the state-of-the-arts on CIFAR-10, CIFAR-100 and ImageNet.
   Top-1 error rates (\%) are reported.
   ResNeXt29: ResNeXt29-2$\times$64d.
   Implementations of all methods are mainly based on~\footnotemark[1] and~\footnotemark[2].
   $^\dagger$: Reported result 29.45$\pm$0.23.
   }
   \label{table:state-of-the-art_cifar_img}
\end{table*}

\subsection{Comparison with the State-of-the-Arts}
\noindent{\bf Competitors.}
We compared PCL with backbone networks (Baseline) and six online KD state-of-the-arts
(DML~\cite{zhang2018deep}, CL~\cite{song2018collaborative}, ONE~\cite{lan2018knowledge},
FFL-S~\cite{kim2019feature}, OKDDip~\cite{chen2020online}, KDCL(-Naive)~\cite{guo2020online}).

\noindent{\bf Setting.}
For fair comparisons, following~\cite{lan2018knowledge},
we used three-branch architectures in compared methods (ONE, CL, FFL-S, OKDDip and PCL)
unless they have to be used with network-based architectures (three parallel networks in DML and KDCL).
Here, although the network-based OKDDip usually yields better performance than the branch-based one,
the former one uses more parameters for training, so we used the branch-based OKDDip.

\noindent{\bf Results.}
As shown in Table~\ref{table:state-of-the-art_cifar_img},
the proposed PCL improves the performance of various backbone networks (baseline)
by approximately 1\% and 2\% on CIFAR-10 and CIFAR-100, respectively.
This shows the effectiveness of PCL for improving the generalisation
performance of various backbone networks.
On CIFAR-10 and CIFAR-100,
PCL achieves the best top-1 error rates compared with the state-of-the-art
online distillation methods.
For example, on CIFAR-10, PCL improves the state-of-the-arts by approximately 0.1\% and 0.3\%
with ResNet-32 and ResNet-110, respectively; Whilst on CIFAR-100,
PCL improves the state-of-the-arts by about 0.3\% and 1.1\% with ResNet-32 and ResNet-110, respectively.
These improvements attribute to the integration of the peer mean teacher and the peer ensemble teacher
into a unified framework.
When extended to the large-scale ImageNet benchmark, 
as shown in Table~\ref{table:state-of-the-art_cifar_img},
PCL improves the baseline by approximately 0.9\%
with ResNet-18.
Compared with the state-of-the-art alternative methods,
PCL still achieves competitive top-1 error rate (about 29.6\% with ResNet-18).

\begin{table}[t]
\begin{center}
\small
    \begin{tabular}{c|p{0.61\columnwidth}|c}
    \hline
    &\multicolumn{1}{c|}{Component}& CIFAR-100\\
    \hline\hline
    &Backbone&23.79$\pm$0.57\\
    \hline
    \multirow{4}{*}{\rotatebox{90}{Proposed}}
    &Backbone+$L_{ce}^p$&23.56$\pm$0.50\\
    &Backbone+$L_{ce}^p$+$L_{ce}^t$&23.48$\pm$0.99\\
    &Backbone+$L_{ce}^p$+$L_{ce}^t$+$L_{pe}$&21.19$\pm$0.62\\ 
    &Backbone+$L_{ce}^p$+$L_{ce}^t$+$L_{pe}$+$L_{pm}$ (full)&{\bf 20.02$\pm$0.55}\\
    \hline
    \multirow{4}{*}{\rotatebox{90}{Variant}}
    &P.E.+Mutual Distillation&21.09$\pm$0.18\\
    &P.E.+Traditional Mean Model(weighted)&21.36$\pm$0.73\\
    &Logit Sum + P.M.&20.43$\pm$0.71\\
    &P.E. + P.M. (full model) &{\bf 20.02$\pm$0.55}\\
    \hline
    \end{tabular}
\end{center}
\caption{Component effectiveness evaluation with ResNet-110 on CIFAR-100.
Top-1 error rates (\%).
P.E.: Peer Ensemble teacher.
P.M.: Peer Mean teacher.
}
\label{table:Component}
\end{table}

\begin{figure}[t]
\begin{center}
   \includegraphics[width=0.75\linewidth]{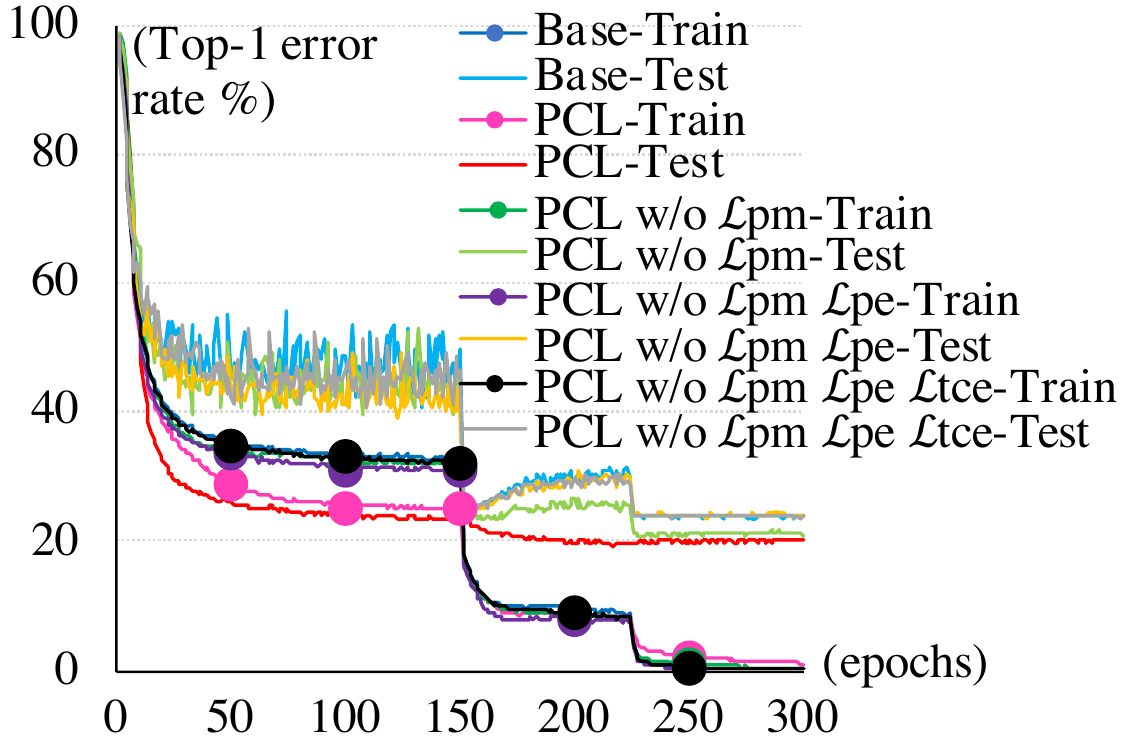}
\end{center}
   \caption{Component effectiveness comparison during training and testing with ResNet-110 on CIFAR-100.
   }
   \label{fig:curve_component}
\end{figure}

\noindent{\bf Discussion.}
These results show the performance advantages of PCL for online KD.
Note that with the peer ensemble teacher and the peer mean teachers,
PCL requires extra computational cost but:
(1) the inference cost of PCL is the same as the target backbone because
we only use a single peer mean model as the target model for test;
(2) the peer mean teachers are updated with Eq.(\ref{eq:mean_teacher}) without the need to perform
backpropagation~\cite{tarvainen2017mean};
(3) the peer ensemble teacher is a multi-branch model and the extra computational cost
is mainly to train the additional classifier.

\footnotetext[1]{\url{https://github.com/DefangChen/OKDDip-AAAI2020}}
\footnotetext[2]{\url{https://github.com/Lan1991Xu/ONE_NeurIPS2018}}

\subsection{Component Effectiveness Evaluation}
From Table~\ref{table:Component}, we can see that:
(1) With all components,
PCL (full model) achieves the best result, demonstrating the effectiveness of integrating of peer
ensemble teacher and peer mean teacher into a unified framework for online KD.
(2) Backbone+$L_{ce}^p$+$L_{ce}^t$+$L_{pe}$
significantly improves the performance of Backbone by about 2.6\%,
showing the effectiveness of the peer ensemble teacher.
(3) PCL (full model) improves Backbone+$L_{ce}^p$+$L_{ce}^t$+$L_{pe}$ by about 1.1\%,
which indicates the effectiveness of the peer mean teacher.
(4) Replacing P.E. or P.M. with some contemporary variants causes performance degradation,
which demonstrates the superiority of the proposed PCL.
Besides, from Fig.~\ref{fig:curve_component}, we can see that PCL with all components (red curve)
gets better generalisation.
Interestingly, the test top-1 error rate (red curve) of PCL (full model)
drops rapidly from 0 to 50 epochs, and then gradually reaches to the optimal value;
In contrast, other methods (w/o $L_{pm}$) fluctuate dramatically, especially from 0 to 225 epochs.
This shows the importance of the peer mean teacher for
learning richer knowledge among peers and optimising a more stable model.

\subsection{Ensemble Effectiveness Evaluation}
We compare our PCL-E with three online KD ensembles:
ONE-E (ensemble of all branches)~\cite{lan2018knowledge}, FFL (FFL-S with fused ensembles)~\cite{kim2019feature},
and OKDDip-E (ensemble of peers)~\cite{chen2020online}.
As shown in Table~\ref{table:Ensembling}, PCL-E improves the state-of-the-arts by
about 0.3\% and 0.6\% on CIFAR-10 and CIFAR-100, respectively.
Besides, compared with ONE-E (the alternative method with the fewest model parameters),
PCL-E achieves significantly better performance but
only increases the number of model parameters by 0.01M and 0.08M with ResNet-110 on
CIFAR-10 and CIFAR-100, respectively.

\begin{table}[t]
\begin{center}
\small
    \begin{tabular}{c|cc|cc}
    \hline
    \multicolumn{1}{c|}{\multirow{2}{*}{Method}} & \multicolumn{2}{c|}{\multirow{1}{*}{CIFAR-10}}
    &\multicolumn{2}{c}{\multirow{1}{*}{CIFAR-100}}\\
    &Top-1&Param.&Top-1&Param.\\
    \hline\hline
    ONE-E&4.75$\pm$0.27&{\bf 2.89M}&20.10$\pm$0.24&{\bf 2.96M}\\
    FFL (fused)&4.99$\pm$0.07&3.10M&21.78$\pm$0.28&3.19M\\
    OKDDip-E&4.79$\pm$0.12&2.91M&20.93$\pm$0.57&2.98M\\
    PCL-E(ours)&{\bf 4.42$\pm$0.12}&2.90M&{\bf 19.49$\pm$0.49}&3.04M\\
    \hline
    \end{tabular}
\end{center}
\caption{Ensemble effectiveness evaluation with ResNet-110 on CIFAR-10/100.
   Top-1 error rates (\%) and the number of model parameters are reported.
}
\label{table:Ensembling}
\end{table}

\begin{figure}
\begin{center}
   \includegraphics[width=0.85\linewidth]{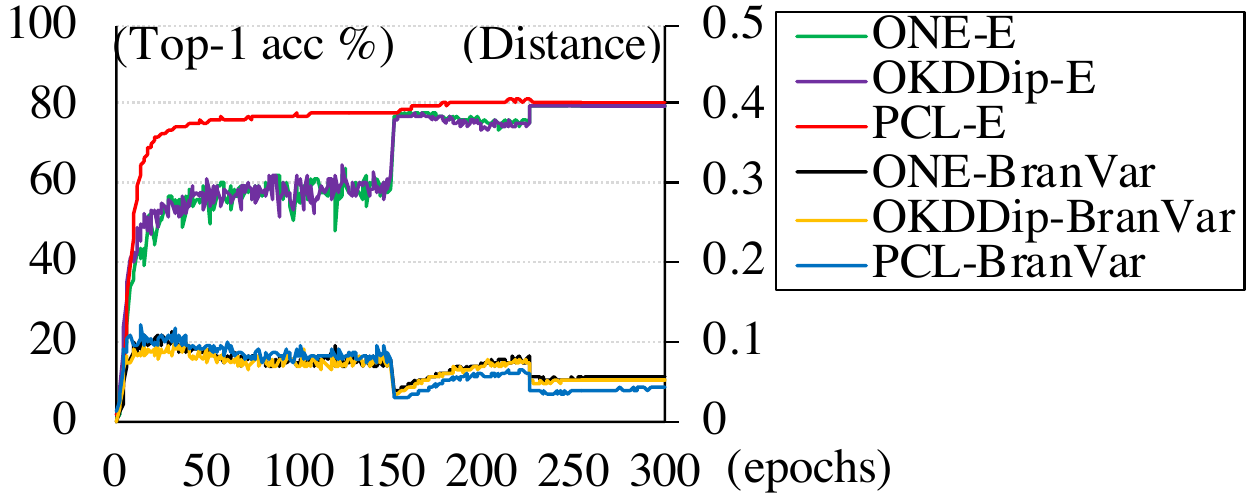}
\end{center}
   \caption{Peer variance for online ensembling analysis with ResNet-110 on CIFAR-100.
   `-BranVar': the branch variance.
   Here, we use top-1 accuracy for better visualisation.
   }
   \label{fig:curve_ensemble_diverity}
\end{figure}

\subsection{Peer Variance for Online Ensembling Analysis}
In Fig.~\ref{fig:curve_ensemble_diverity}, we analyse the peer (branch) variance for online ensembling
over the training epochs.
Here, we computed the average Euclidean distance between the predictions of every two branches as the branch diversity
and used the average diversity of $m$ branches as the branch variance.
From Fig.~\ref{fig:curve_ensemble_diverity}, we can see that:
(1) From 0 to 150 epochs, the top-1 accuracy of PCL-E soars to a high level outperforming other methods,
and meanwhile, the branch variance of PCL (PCL-BranVar) is larger than other methods.
This indicates that at the early stage, although the generalisation capability of the model is poor,
each branch in PCL collaborates better to facilitate online knowledge distillation.
(2) From 150 to 300 epochs, the top-1 accuracy of PCL-E is still better than the alternatives,
whilst the branch variance of PCL becomes smaller than the alternatives.
The main reason is that at this stage,
the generalisation of each peer is significantly improved
and the temporally aggregated model of each peer becomes stable (with accurate and similar predictions).
This also results in a stronger ensemble model (see Table~\ref{table:Ensembling})
and a more generalised target model (see Table~\ref{table:state-of-the-art_cifar_img}).

\subsection{Further Analysis and Discussion}

\begin{table}[t]
\begin{center}
\small
    \begin{tabular}{c|ccc}
    \hline  
    Dataset&Baseline&KD$^\dagger$&PCL\\
    \hline\hline
    CIFAR-10&6.74$\pm$0.15&5.82$\pm$0.12&{\bf 5.67$\pm$0.12}\\
    CIFAR-100&28.72$\pm$0.19&26.23$\pm$0.21&{\bf 25.86$\pm$0.16}\\
    \hline
    \end{tabular}
\end{center}
\caption{Comparison with two-stage distillation with ResNet-32 on CIFAR-10/100.
   Top-1 error rates (\%).
   $^\dagger$: Use ResNet-110 as the teacher model.
}
\label{table:Two-stage}
\end{table}

\noindent{\bf Comparison with Two-Stage Distillation.}
In Table~\ref{table:Two-stage}, we compare PCL with the traditional two-stage KD~\cite{hinton2015distilling}.
We can see that although PCL does not pre-train a high-capacity teacher model (\eg ResNet-110),
it still achieves better performance than the two-stage KD.
This attributes to the integration of the peer ensemble teacher and the peer mean teacher into
a unified framework for online knowledge distillation.

\begin{figure}[t]
\begin{center}
   \includegraphics[width=0.99\linewidth]{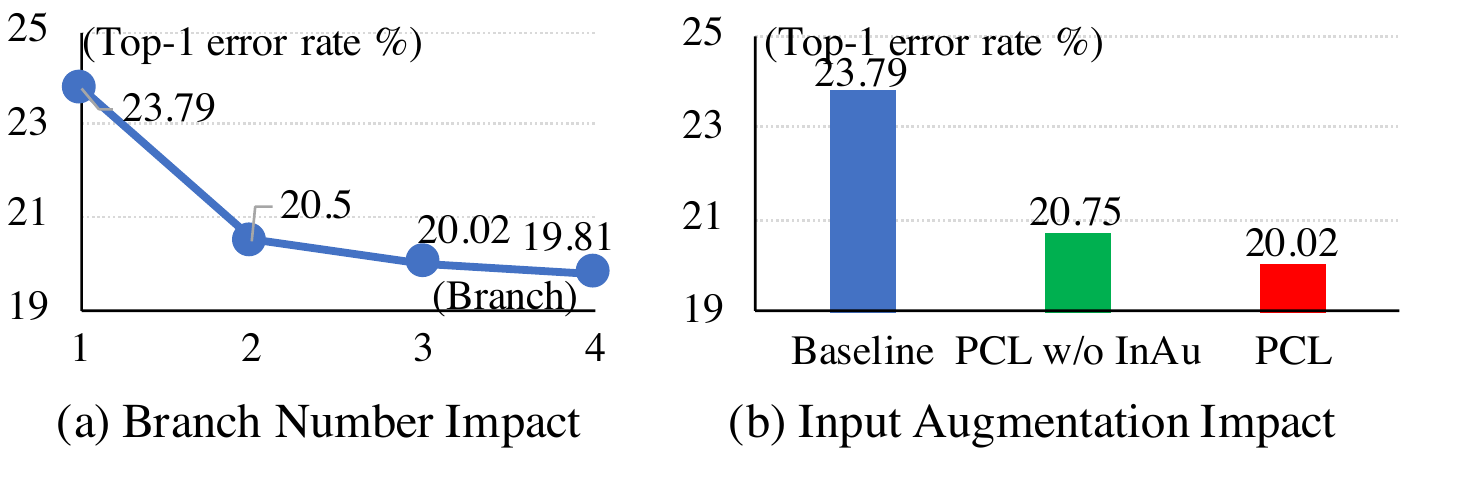}
\end{center}
   \caption{Evaluating the impact of (a) different number of branches and (b) input augmentation for PCL
   with ResNet-110 on CIFAR-100.
   }
   \label{fig:Branch_InputAu}
\end{figure}

\noindent{\bf Branch Number.}
As shown in Fig.~\ref{fig:Branch_InputAu}(a),
the performance of PCL improves when more branches are used.
In the four-branch setting, PCL (19.8\%) still performs competitively against OKDDip
(21.1\% as reported in~\cite{chen2020online}).

\noindent{\bf Input Augmentation.}
As shown in Fig.~\ref{fig:Branch_InputAu}(b),
without using multiple input augmentation (PCL w/o InAu), the performance of PCL decreases
(by approximately 0.7\%), but it still achieves compelling performance.
This further verifies the effectiveness of the model design in PCL.

\section{Conclusion}
In this work, we presented a novel Peer Collaborative Learning (PCL) method for
online knowledge distillation,
which integrates online ensembling and network collaboration into a unified framework.
We assembled feature representations from peers as
the online high-capacity peer ensemble teacher
and used the temporal mean model of each peer as the peer mean teacher.
Doing so allows improving the quality of online knowledge distillation in
a one-stage end-to-end trainable fashion.
Extensive experiments with a variety of backbone networks
show the superiority of the proposed method over the state-of-the-art
methods on CIFAR-10, CIFAR-100 and ImageNet.

\section{Acknowledgements}
This work is supported by Vision Semantics Limited,
Alan Turing Institute Turing Fellowship, and Innovate UK Industrial Challenge Project on Developing and Commercialising
Intelligent Video Analytics Solutions for Public Safety (98111-571149), Queen Mary University of London Principal’s Scholarship.

\bibliography{WuEtAl_OnlineDistil}
\end{document}